# Exploiting Shape Cues for Weakly Supervised Semantic Segmentation

Sungpil Kho[a,*], Pilhyeon Lee[b,*], Wonyoung Lee[a], Minsong Ki[c,**] and Hyeran Byun[a,b,***]

[a]Graduate School of Artificial Intelligence, Yonsei University, Seoul, Republic of Korea
[b]Department of Computer Science, Yonsei University, Seoul, Republic of Korea
[c]AI Imaging Tech. Team, LG Uplus, Seoul, Republic of Korea

## ABSTRACT

Weakly supervised semantic segmentation (WSSS) aims to produce pixel-wise class predictions with only image-level labels for training. To this end, previous methods adopt the common pipeline: they generate pseudo masks from class activation maps (CAMs) and use such masks to supervise segmentation networks. However, it is challenging to derive comprehensive pseudo masks that cover the whole extent of objects due to the local property of CAMs, *i.e.*, they tend to focus solely on small discriminative object parts. In this paper, we associate the locality of CAMs with the texture-biased property of convolutional neural networks (CNNs). Accordingly, we propose to exploit shape information to supplement the texture-biased CNN features, thereby encouraging mask predictions to be not only comprehensive but also well-aligned with object boundaries. We further refine the predictions in an online fashion with a novel refinement method that takes into account both the class and the color affinities, in order to generate reliable pseudo masks to supervise the model. Importantly, our model is end-to-end trained within a single-stage framework and therefore efficient in terms of the training cost. Through extensive experiments on PASCAL VOC 2012, we validate the effectiveness of our method in producing precise and shape-aligned segmentation results. Specifically, our model surpasses the existing state-of-the-art single-stage approaches by large margins. What is more, it also achieves a new state-of-the-art performance over multi-stage approaches, when adopted in a simple two-stage pipeline without bells and whistles.

## 1. Introduction

The goal of semantic segmentation is to predict pixel-level categories in a given image. Thanks to the various applications, such as scene understanding [1], autonomous driving [2], and image editing [3], it has been actively studied by the research community. Particularly, a number of fully supervised methods are devised for the task and achieve excellent segmentation performances [4, 5, 6, 7, 8]. Nevertheless, the expensive cost for obtaining pixel-wise annotations largely limits their scalability and practicability.

To mitigate the cost issue, utilizing weak supervision has received increasing attention recently, mainly image-level tags [9, 10, 11, 12]. Given only image-level labels, existing methods employ class activation maps (CAMs)[1] [13] as an initial seed region and derive pseudo masks from it [14, 10]. However, with only the image-level classification loss, CAMs tend to highlight only small discriminative parts rather than the full extent of an object [15, 16, 17]. This locality of CAMs leads the resulting pseudo masks to be less comprehensive as well as to disagree with object outlines. To compensate for it, existing multi-stage methods try to refine CAMs in an *offline* fashion and utilize them to train an external segmentation network [10, 18]. Nevertheless, they necessitate high levels of computational complexity and a long training time, which eventually compromises the advantage of weak supervision in terms of costs [19].

In this paper, we aim to generate improved CAMs *on the fly* that better cover object regions. To that end, we first make a connection between the locality of CAMs and the texture bias of convolutional neural networks (CNNs). As diagnosed in the recent literature [20, 21, 22], CNNs are heavily biased toward texture information when classifying an object. Therefore, it is reasonable that CAMs show high activation only for the discriminative texture patterns of small object parts (*e.g.*, the faces of cats), as demonstrated in Fig. 1c. Conversely, one may imagine a shape-biased CNN [20] – its CAM would focus solely on the object boundaries (*e.g.*, the outlines of cats). To sum up, both types of biases are complementary and indispensable for obtaining comprehensive CAMs that can capture object regions to a whole extent. This motivates us to explore leveraging shape-biased features together with the conventional texture-biased ones of CNNs so as to generate better CAMs.

From this motivation, we introduce a novel framework for weakly supervised semantic segmentation, where shape information plays a role as *shape cues* in producing complete CAMs that localize the entire regions of objects. Specifically, we design a shape cue module (SCM) that learns shape-related features by intentionally removing texture information. The shape-biased features extracted by SCM are injected back to the network, in order to supplement the original texture-biased features during the decoding process. Consequently, the model can derive more comprehensive segmentation maps that not only cover the discriminative parts but align well with object boundaries. Moreover, to efficiently generate precise pseudo masks during training, we propose an *online* mask refinement method, namely semantics-augmented pixel refinement (SPR). Technically, our SPR takes into account class semantics as well as color information to compute local inter-pixel affinities. Based on the affinities, it can effectively refine the initial predictions, resulting in high-quality pseudo masks. They are in turn used to supervise our model in an *end-to-end* fashion.

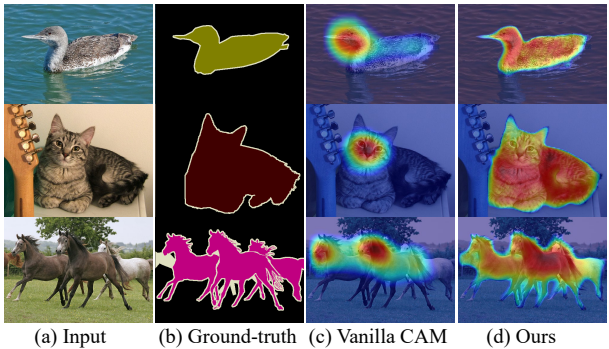

| (a) Input | (b) Ground-truth | (c) Vanilla CAM | (d) Ours |

**Figure 1:** Visualization of the effectiveness of our method. Compared to the vanilla CAM that pays attention to only local parts (*e.g.*, faces), our method is able to produce more comprehensive activation maps by exploiting object shape cues.

With the help of the shape cues and the effective mask refinement, our model is able to produce comprehensive activation maps with well-aligned boundaries, as shown in Fig. 1d. Through extensive experiments on PASCAL VOC 2012 [23], we clearly validate the effectiveness of the proposed methods in improving the segmentation performance in the weakly supervised setting. Moreover, we demonstrate the superiority of our model over existing state-of-the-art single-stage approaches in terms of mask quality and boundary alignment. Our model also sets a new state-of-the-art in the multi-stage settings when adopted in a *simple* two-stage pipeline without bells and whistles.

To summarize, our contributions are four-fold.

- We shed light on the connection between the locality of CAMs and the texture bias of CNNs, which has hardly been handled before.

- We introduce a novel weakly supervised segmentation method that explicitly leverages shape-biased features as *shape cues* for producing comprehensive segmentation maps, overcoming the locality of CAMs.

- We propose a new pseudo mask generation method, where both color and semantic information are leveraged for obtaining pairwise pixel affinities, thereby accurately refining initial mask predictions.

- On PASCAL VOC 2012, the most popular benchmark, our method achieves a new state-of-the-art performance with a significant margin in both the single and the multi-stage settings.

## 2. Related Work

### 2.1. Fully Supervised Semantic Segmentation

The goal of semantic segmentation is to classify each pixel in a given image. A majority of fully supervised methods are built upon the encoder-decoder architecture to predict segmentation masks from latent representation [24, 25, 6]. Meanwhile, some methods adopt the attention mechanism to capture rich contextual relations [26, 27, 28, 29]. In addition, there appear attempts to find the optimal model design by neural architecture search [30, 31]. Recently, researchers bring the remarkable success of Transformers [32] to the segmentation task [7, 33, 8]. Despite the great performances, fully supervised methods suffer from the high labeling cost, which triggers researchers to explore weak supervision [9] or even the unsupervised setting [34, 35].

### 2.2. Weakly Supervised Semantic Segmentation

Weakly supervised learning aims to mitigate heavy annotation costs and has widely been studied in image [13, 36] and video domains [37, 38, 39] using various label forms. Specifically, most of the weakly supervised semantic segmentation work utilizes image-level labels owing to the cheapest cost, while a variety of supervision levels have also been employed, such as points [40], scribbles [41, 42], and bounding boxes [43, 44, 45, 46]. Our framework employs image-level labels for training, and existing methods can be divided into two mainstreams.

**Multi-stage approaches** first obtain class activation maps (CAMs) [13] using a classification network and then refine them in an *offline* manner, which are later used as pseudo labels to train external segmentation models. Early approaches examine the adjacent pixels of initial confident seeds to grow seed regions [47, 9]. Moreover, several methods consider broader pixel affinities using the random walk [10] or the self-attention [11]. There are also some attempts to improve the initial seed quality by the stochastic convolution [15] or the online attention accumulation [48]. CONTA [14] adopts the backdoor adjustment to alleviate confounding biases, while CDA [49] designs a copy-and-paste augmentation to decouple the context from objects. Recent methods mainly focus on discovering less discriminative object parts by erasing the dominant regions [50, 51], using complementary image patches [52], or pushing images away from the decision boundary [16]. Meanwhile, some methods attempt to refine segmentation results by exploiting saliency maps as post-processing [53, 54] or auxiliary supervision [55, 56] In addition, several recent methods consider the inter-image relation to capture richer information [57, 58, 59].

Among the prior work, there are several methods that try to refine the initial segmentation results to be aligned with boundaries. The first group relies on the saliency labels or predictions [54, 55, 58]. However, the saliency maps require extra human annotations and/or auxiliary model training. Also, they intrinsically highlight only the dominant object in the scene and ignore small objects, which can be detrimental to the segmentation task where both types of objects need to be captured. The second group tries to estimate semantic boundaries. Amongst, some methods [10, 18] take an indirect way by allowing a model to learn inter-pixel affinities, while another approach [60] directly trains a boundary detection model by generating pseudo boundary labels. Nonetheless, they all need auxiliary networks for the purpose and cannot be trained in an end-to-end fashion, *i.e.*, they require training of another segmentation model in the next stage. In contrast to them, our method (a) does not require any saliency labels or predicted maps, (b) does not need any auxiliary networks, and (c) is seamlessly integrated into a single-stage segmentation pipeline and provides shape cues in an end-to-end fashion.

**Single-stage approaches** streamline the complicated training of multi-stage counterparts in order to save the expensive training cost. EM-Adapt [61] introduces an online expectation-maximization method under weak annotation constraints. CRF-RNN [62] fuses top-down and bottom-up activation maps with smoothness visual cues, while RRM [19] integrates the classification and the segmentation networks into an end-to-end framework. Moreover, SSSS [12] introduces three desirable concepts for precise segmentation and improves the segmentation quality based on them. Most recently, AA&LR [63] proposes the adaptive affinity loss and the label reassigning strategy to better learn from noisy pseudo masks.

Our method follows the efficient single-stage pipeline. Different from previous work, we shed light on the connection between the locality of CAMs and the texture biases of CNNs. Accordingly, we propose to explicitly leverage *shape cues*, thereby allowing the predicted masks to capture the objects accurately.

## 2.3. Texture Biases in Convolutional Neural Networks

In the recent literature [20, 64], it is demonstrated that, unlike humans, convolutional neural networks (CNNs) have strong biases towards textures rather than shapes, *i.e.*, they heavily rely on texture information when classifying objects. It is also revealed that shape-biased features are more robust than texture-biased counterparts against domain shifts [65] and adversarial attacks [66]. Accordingly, a variety of methods have been proposed to ameliorate the texture biases, such as image stylization [20], information erasure [67], dedicated augmentation schemes [21], adversarial training [68], and separated supervision [22].

In this paper, we argue that the texture-biased property of CNNs hampers the segmentation performance in the weakly supervised setting. To tackle this challenge, we discover shape-related features and use them as *shape cues* for producing more precise segmentation masks.

## 3. Proposed Method

As depicted in Fig. 2, our framework consists of two parts: the shape-aware segmentation network and the self-supervised training with pseudo masks. The shape-aware segmentation network takes an image as input and produces mask predictions. To enhance the mask quality, we introduce a novel shape cue module (SCM) that distills shape-related features from the encoder. During decoding, they are used as *shape cues* to encourage the segmentation mask to be comprehensive. Afterward, our semantics-augmented pixel refinement (SPR) further hones the mask prediction by leveraging both color information and class semantics, leading to accurate pseudo masks. The resulting pseudo masks in turn provide pixel-level supervision to the segmentation network in an online fashion.

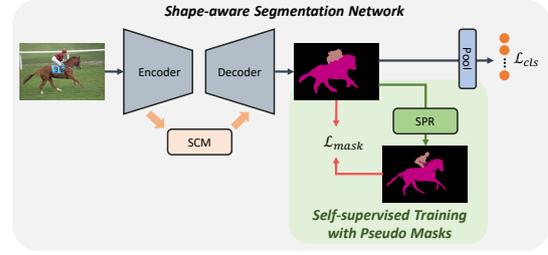

**Figure 2:** Overview of the proposed framework. It consists of the shape-aware segmentation network and the self-supervised training with pseudo masks.

### 3.1. Shape-aware Segmentation Network
#### 3.1.1. Overview

Our shape-aware segmentation network follows the architectural design of DeepLabV3 [5]. An input image $X \in \mathbb{R}^{H \times W \times 3}$ goes through the encoder, where $H$ and $W$ are the height and the width of the image, respectively. Taking the encoded feature as input, the decoder predicts an initial segmentation score map. Following the existing work [12], we add an auxiliary map for the background class filled with constant values (*i.e.*, 1). This results in the score map $S \in \mathbb{R}^{h \times w \times (K+1)}$, where $(h, w)$ indicates the size of the score map and $K + 1$ denotes the number of object classes plus the background. To train the model with image-level labels, we obtain the image-level class scores by performing normalized global weighted pooling. In detail, we first obtain the normalized score map $M$, *i.e.*, $m_i^c = \frac{\exp(s_i^c)}{\sum_{c'=1}^{K+1} \exp(s_i^{c'})}$, where $m_i^c$ is the normalized score for class $c$ of the $i$-th pixel[2]. Then, the image-level class score for the $c$-th class is derived by the attended sum of pixel-level scores as follows.

$$v^c = \frac{\sum_{\forall i} m_i^c \cdot s_i^c}{\epsilon + \sum_{\forall i} m_i^c}, \tag{1}$$

where $\epsilon$ is a small constant for numerical stability. Afterward, we apply the sigmoid function on the obtained scores to get the image-level class probabilities, *i.e.*, $\hat{y}^c = \sigma(v^c)$.

Given the image-level class probabilities $\hat{\mathbf{y}}$ and the image-level weak labels $\mathbf{y}$, we calculate the classification

---

[2]For better presentation, we use pixel indices rather than $(x, y)$ coordinates throughout this paper.

loss using the binary cross-entropy as follows.

$$\mathcal{L}_{\text{cls}} = -\frac{1}{K} \sum_{c=1}^{K} \left( y^c \log \hat{y}^c + (1 - y^c) \log(1 - \hat{y}^c) \right). \quad (2)$$

We note that the background class is excluded from the loss calculation.

### 3.1.2. Shape cue module

As discussed in Sec. 1, the CNN-based encoder would learn biased representation toward texture, which results in incomplete masks in the decoding process. To improve the mask quality, we design the shape cue module (SCM) that utilizes shape-related features as shape cues along with the original texture-biased features. Although various methods can be applied to extract shape information from CNNs, such as texture diversification [20] and adversarial training [68], we here instantiate our SCM based on the information erasure approach [67] owing to its simplicity and generality. In a nutshell, we pull out shape-biased features from the encoder by intentionally erasing its texture information.

The next question is, *how can we selectively remove texture information?* Generally, texture regions are known to be less informative than shape ones, since they tend to be highly similar to their neighborhood. Formally, the information contained in a given local region (*i.e.*, patch) can be estimated with the Shannon entropy [69]. However, since directly estimating the distribution of a patch is impractical, we follow Shi et al. [67] to approximate the distribution by treating the neighboring patches as its samples. Consequently, the self-information of the patch **p** can be estimated using the kernel density estimator with the Gaussian kernel as follows.

$$\hat{\mathcal{I}}(\mathbf{p}) = -\log \frac{1}{|\mathcal{B}_{\mathbf{p}}|} \sum_{\forall \mathbf{p}' \in \mathcal{B}_{\mathbf{p}}} \frac{\exp\left(-\|\mathbf{p} - \mathbf{p}'\|^2 / 2h^2\right)}{\sqrt{2\pi}h}, \quad (3)$$

where $\mathcal{B}_{\mathbf{p}}$ is the set of all neighboring patches of **p** within the pre-defined distance $R_1$, $h$ is the bandwidth of the Gaussian kernel, and $|\cdot|$ is the cardinality operator.

Considering a patch placed on a texture region, for example, it is likely to have high color similarity with neighbors (the same texture) and therefore will get low self-information. On the contrary, another patch containing part of an object shape would have high self-information, since they tend to be unique among their neighborhood. Based on this tendency, we remove texture-related features by stochastically zeroing out the neurons of texture regions, where the dropping probability is inversely proportional to the self-information of the regions. Specifically, we define the dropping probability for the center pixel of the patch **p**, *i.e.*, $c_{\mathbf{p}}$, with a Boltzmann distribution as follows.

$$r(c_{\mathbf{p}}) \propto e^{-\hat{\mathcal{I}}(\mathbf{p})/\tau}, \quad (4)$$

where $\tau$ is a temperature parameter that adjusts the smoothness of dropping probabilities. For instance, a small $\tau$ leads to a sharp probability distribution.

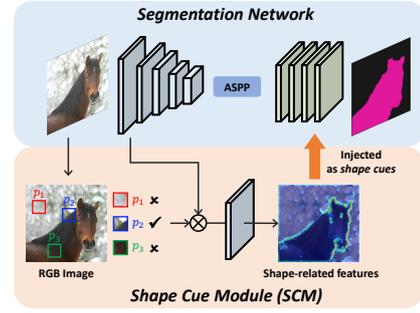

**Figure 3:** Details of the proposed shape cue module (SCM). Our SCM extracts shape-biased features and allows the decoder to use them as *shape cues* for accurate segmentation.

We obtain the dropping probability for every pixel, based on which we block out the input feature map. For example in Fig. 3, the green and the red boxes have been dropped due to the low information, whereas the blue box survives and passes through the following convolutional layer to acquire the texture-dropped (*i.e.*, shape-biased) features. We then feed the obtained shape-related features into the decoder by concatenating them with intermediate features. By utilizing them as *shape cues*, our segmentation network can effectively produce comprehensive and boundary-aligned segmentation masks, as verified in Sec. 4.

### 3.2. Self-supervised Training with Pseudo Masks

To improve the segmentation performance, existing weakly supervised segmentation methods typically generate pseudo mask labels in either an online (single-stage) [19, 12] or an offline manner (multi-stage) [9, 10]. Our method lies in line with the single-stage approaches, and therefore obtains pseudo masks *on the fly* by efficiently refining the initial prediction.

### 3.2.1. Semantics-augmented pixel refinement

Existing single-stage approaches mainly exploit the color information of images (*i.e.*, RGB) to refine initial predictions by performing dense-CRF [70] or using a local RGB affinity kernel [12]. However, hinging solely on the color space might be sub-optimal for mask refinement. For instance, considering a tree, it is implausible to propagate the predictions of leaves into branches based on color information, although they constitute the same object. To tackle this challenge, we propose to consider class semantics as well as color information to compute the pixel affinities. Specifically, we introduce a semantics-augmented pixel refinement (SPR), where two different kernels are employed respectively for color and label spaces. Formally, the joint kernel function between the pixels $i$ and $j$ can be formulated for class $c$ as follows.

$$k^c(i, j) = -\alpha \frac{\|\mathbf{x}_i - \mathbf{x}_j\|}{\sigma_{\mathbf{x}}^2} - (1 - \alpha) \frac{|s_i^c - s_j^c|}{(\sigma_s^c)^2}, \quad (5)$$

where $\alpha$ is a hyper-parameter to balance two types of affinities, while $\mathbf{x}_i$ and $s_i^c$ indicate the pixel intensity and the score for class $c$ of the $i$-th pixel respectively, with their standard deviations of $\sigma_{\mathbf{x}}$ and $\sigma_s^c$.

With the defined kernel in Eq. 5, we get the local affinities for pixel $i$ by applying the softmax within a local window centered at it, *i.e.*, $\delta_{j\to i}^c = \frac{\exp(k^c(i,j))}{\sum_{\forall j' \in \mathcal{N}_i} \exp(k^c(i,j'))}$, where $\mathcal{N}_i$ is the set of pixels in the window with its radius of $R_2$. Thereafter, the normalized score map $M$ is repeatedly refined using the local affinities. For example, the refined score for class $c$ of the pixel $i$ at the $t$-th iteration can be obtained by:

$$m_{i;t}^c = \sum_{\forall j \in \mathcal{N}_i} \delta_{j\to i}^c \cdot m_{j;t-1}^c, \tag{6}$$

where $\mathcal{N}_i$ is the set of pixels within the distance $R_2$, and the beginning value ($t = 0$) is set to the initial score, *i.e.*, $m_{i;0}^c = m_i^c$. The refinement is performed for all pixels and the object and the background classes present in the image[3]. The refining process is at the $t$-th iteration is illustrated in Fig. 4.

After a total of $T$ times refinement ($T = 10$), we get the final scores $m_{i;T}^c$ for class $c$. Then, we generate the pseudo mask $\tilde{M}$ by selecting the class with the highest probability and thresholding on $\mathbf{m}_{i;T}$ with the pre-defined threshold $\theta$. We assign one-hot pseudo labels only to the remaining pixels, which constitute the valid set $\mathcal{V}$.

We note that our SPR can be viewed as a generalization of the existing RGB-based refinement methods [19, 12]. For instance, when $\alpha = 1$, our SPR relies exclusively on pixel intensities, which is what the previous methods do. However, working only on pixel intensities would be sub-optimal for mask refinement and the class semantics indeed helps to improve the pseudo mask quality, which we validate in Sec. 4. Meanwhile, our SPR also relates to AffinityNet [10] in that both consider class semantics for mask refinement. Nonetheless, they clearly differ from each other in the following aspects. AffinityNet requires training of an auxiliary network and performs refinement in an *offline* manner. In contrast, our SPR is seamlessly integrated in an *end-to-end* framework and efficiently refines initial masks in an *online* fashion without any external network.

### 3.2.2. Learning from pseudo masks

With the pseudo masks, we train the model to produce similar segmentation results to them. From this point of view, we compute the conventional pixel-wise loss between predictions and pseudo masks. In addition, we adopt the region-wise loss that optimizes the segmentation results at the region level.

Firstly, the *pixel-wise loss* computes the cross-entropy losses for individual pixels and aggregates them as follows.

$$\mathcal{L}_{\text{pixel}} = -\frac{1}{|\mathcal{V}|} \sum_{\forall i \in \mathcal{V}} \sum_{c=1}^{K+1} \tilde{m}_i^c \log m_i^c, \tag{7}$$

where $m_i^c$ and $\tilde{m}_i^c$ are the prediction and the pseudo GT of pixel $i$ for class $c$, respectively. We compute the loss for only the pixels in the valid set $\mathcal{V}$ and ignore the other unconfident pixels. We also normalize the class-wise loss to handle the class imbalance.

---

[3]In practice, the refinement is implemented with the GPU-friendly convolution operator, thus is lightweight in terms of costs.

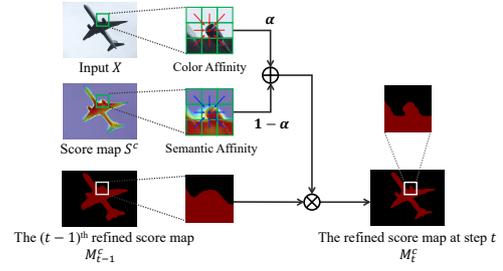

**Figure 4:** Illustration of semantics-augmented pixel refinement (SPR). Our SPR considers class semantics as well as pixel intensities to compute the inter-pixel affinities between the center and the neighboring pixels. Note that the predicted segmentation maps, *i.e.*, $arg\,max(M_i)$, are presented to represent the refined score maps for the illustrative purpose.

Since the pixel-wise loss penalizes the pixel-level predictions independently, it ignores the inter-pixel relation that is informative for segmentation. Therefore, along with the pixel-wise loss, we also consider the region-wise supervision. Motivated by Ke et al. [71], we design a loss function that involves structural information of pixels. Technically, we first define a local window with its radius $R_3$ centering at the $i$-th pixel in the pseudo masks. Afterward, we separate the neighboring pixels in the window, $\mathcal{M}_i$, into two sets according to whether they agree with the center point $i$. If a pixel $j \in \mathcal{M}_i$ produces a different class prediction with the center $i$, *i.e.*, $arg\,max(\tilde{\mathbf{m}}_j) \neq arg\,max(\tilde{\mathbf{m}}_i)$, it is considered a boundary point and inserted into the boundary set $\mathcal{M}_i^{\text{bnd}}$. On the contrary, we put a pixel into the non-boundary set $\mathcal{M}_i^{\text{non}}$, if it shares the same class prediction with the center $i$. We note that every pixel in the window belongs to exactly one of the subsets.

With the two disjoint subsets, we compute the *region-wise loss*. Concretely, we encourage the class probability distribution of the center pixel $i$ to be similar to those of pixels in the non-boundary set, while repelling those of pixels in the boundary set. The loss function is formulated as follows.

$$\mathcal{L}_{\text{region}} = -\frac{1}{|\mathcal{V}|} \sum_{\forall i \in \mathcal{V}} \frac{1}{|\mathcal{M}_i|} \sum_{\forall j \in \mathcal{M}_i} \ell_{i,j},$$

where $\ell_{i,j} = \begin{cases} D_{KL}(\mathbf{m}_j || \mathbf{m}_i), & \text{if } j \in \mathcal{M}_i^{\text{non}} \\ max(0, \gamma - D_{KL}(\mathbf{m}_j || \mathbf{m}_i)), & \text{if } j \in \mathcal{M}_i^{\text{bnd}} \end{cases},$ (8)

$D_{KL}(\cdot)$ denotes the Kullback-Leibler divergence between two class probability distributions, and $\gamma$ is a margin.

In the weakly supervised setting, the generated pseudo masks are often noisy and unreliable, especially at the early training stage. This makes the region-wise loss misleading. Therefore, we use an annealing scheduling for the region-wise loss to stabilize the training. In summary, the overall loss function from the pseudo masks can be computed by:

$$\mathcal{L}_{\text{mask}} = \mathcal{L}_{\text{pixel}} + \lambda \mathcal{L}_{\text{region}}, \tag{9}$$

where $\lambda$ is a weighting factor for stable training, which steadily increases from 0 to 1 during the training.

### 3.3. Joint Training and Inference

As aforementioned, our model follows the single-stage pipeline and thus is trained in an *end-to-end* manner. The overall training object is the sum of the classification loss (Eq. 2) and the mask loss (Eq. 9).

$$\mathcal{L}_{\text{total}} = \mathcal{L}_{\text{cls}} + \mathcal{L}_{\text{mask}}. \tag{10}$$

For inference, we feed an image into our model and use the mask prediction by the decoder as the final result. Note that the pseudo mask generation is not performed at test time.

## 4. Experiments

### 4.1. Experimental Settings

#### 4.1.1. Dataset and evaluation metrics

For evaluation, we use the most popular benchmark for weakly supervised semantic segmentation: PASCAL VOC 2012 [23]. It contains 20 object categories and includes 1,464, 1,449, and 1,456 images respectively for training, validation, and test. Following the convention, we augment the training set with the additional images provided by SBD [72], leading to 10,582 training samples in total. We measure the mean Intersection-over-Union (mIoU) between predicted masks and ground-truths.

#### 4.1.2. Implementation details

Our segmentation network strictly follows the structure of DeepLabV3 [5], an encoder-decoder architecture with atrous spatial pyramid pooling with its output stride of 16. For a fair comparison, we use WideResNet-38 [73] pre-trained on ImageNet [74] as the encoder and adopt a four-layer decoder to predict segmentation masks. Our model is trained in an end-to-end manner for 20 epochs using the SGD optimizer with the weight decay of $10^{-4}$ and the momentum of 0.9. The initial learning rate is set to $10^{-3}$ for the backbone classification network and $10^{-2}$ for the other modules. We follow the learning rate decay strategy of SEAM [11] with the decay rate of 0.9.

#### 4.1.3. Hyperparameter settings

**Data augmentation.** Following the convention [10, 75], we augment each input training image using the following three strategies: (1) random scaling with the ratio range of [0.9, 1.0], (2) random cropping with the resolution of $448 \times 448$, and (3) horizontal flip with the probability of 0.5.

**Shape cue module.** To estimate the self-information of each patch, we use randomly sampled 9 patches from the neighborhood within the manhattan radius $R_1 = 7$. For the Gaussian kernel, we set its bandwidth $h$ to 1. After obtaining the self-information, we derive the dropping probability using the Boltzmann distribution with the smoothness parameter $\tau$ of 0.5 followed by normalization.

| Method | Backbone | Sup. | CRF | val | test |
|---|---|---|---|---|---|
| DeepLabV3 [5] | ResNet-101 | $\mathcal{F}$ | ✗ | - | 85.7 |
| WideResNet-38 [73] | | $\mathcal{F}$ | ✗ | 80.8 | 82.5 |
| **Multi-stage** | | | | | |
| DSRG [9] CVPR'18 | ResNet-101 | $\mathcal{I}, \mathcal{S}$ | ✓ | 61.4 | 63.2 |
| FickleNet [15] CVPR'19 | ResNet-101 | $\mathcal{I}, \mathcal{S}$ | ✓ | 64.9 | 65.3 |
| OAA [48] ICCV'19 | ResNet-101 | $\mathcal{I}, \mathcal{S}$ | ✓ | 65.2 | 66.4 |
| CIAN [53] AAAI'20 | ResNet-101 | $\mathcal{I}, \mathcal{S}$ | ✓ | 64.1 | 64.7 |
| ICD [57] CVPR'20 | ResNet-101 | $\mathcal{I}, \mathcal{S}$ | ✓ | 67.8 | 68.0 |
| LSISU [76] PR'21 | ResNet-101 | $\mathcal{I}, \mathcal{S}$ | ✓ | 68.4 | 68.9 |
| NSROM [54] CVPR'21 | ResNet-101 | $\mathcal{I}, \mathcal{S}$ | ✓ | 68.3 | 68.5 |
| EDAM [58] CVPR'21 | ResNet-101 | $\mathcal{I}, \mathcal{S}$ | ✓ | 70.9 | 70.6 |
| EPS [55] CVPR'21 | ResNet-101 | $\mathcal{I}, \mathcal{S}$ | ✓ | 71.0 | 71.8 |
| AuxSegNet [16] ICCV'21 | ResNet-101 | $\mathcal{I}, \mathcal{S}$ | ✓ | 68.1 | 68.0 |
| AffinityNet [10] CVPR'18 | WideResNet-38 | $\mathcal{I}$ | ✓ | 61.7 | 63.7 |
| IRN [18] CVPR'19 | ResNet-50 | $\mathcal{I}$ | ✓ | 63.5 | 64.8 |
| SEAM [11] CVPR'20 | WideResNet-38 | $\mathcal{I}$ | ✓ | 64.5 | 65.7 |
| ICD [57] CVPR'20 | ResNet-101 | $\mathcal{I}$ | ✓ | 64.1 | 64.3 |
| BES [60] ECCV'20 | ResNet-101 | $\mathcal{I}$ | ✓ | 65.7 | 66.6 |
| MCIS [59] ICCV'20 | ResNet-101 | $\mathcal{I}$ | ✓ | 66.2 | 66.9 |
| CONTA [14] Neurips'20 | WideResNet-38 | $\mathcal{I}$ | ✓ | 66.1 | 66.7 |
| AdvCAM [16] CVPR'21 | ResNet-101 | $\mathcal{I}$ | ✓ | 68.1 | 68.0 |
| ECS-Net [50] ICCV'21 | WideResNet-38 | $\mathcal{I}$ | ✓ | 66.6 | 67.6 |
| CDA [49] ICCV'21 | WideResNet-38 | $\mathcal{I}$ | ✓ | 66.1 | 66.8 |
| CGNet [51] ICCV'21 | WideResNet-38 | $\mathcal{I}$ | ✓ | 68.4 | 68.2 |
| CPN [52] ICCV'21 | ResNet-101 | $\mathcal{I}$ | ✓ | 67.8 | 68.5 |
| RIB [17] Neurips'21 | ResNet-101 | $\mathcal{I}$ | ✓ | 68.3 | 68.6 |
| Ours (two-stage) | WideResNet-38 | $\mathcal{I}$ | ✗ | 67.9 | 68.6 |
| | | | ✓ | **69.5** | **70.5** |
| **Single-stage** | | | | | |
| EM-Adapt [61] ICCV'15 | VGG16 | $\mathcal{I}$ | ✓ | 38.2 | 39.6 |
| SEC [47] ECCV'16 | VGG16 | $\mathcal{I}, \mathcal{S}, \mathcal{D}$ | ✓ | 50.7 | 51.7 |
| CRF-RNN [62] CVPR'17 | VGG16 | $\mathcal{I}$ | ✓ | 52.8 | 53.7 |
| RRM [19] AAAI'20 | WideResNet-38 | $\mathcal{I}$ | ✓ | 62.6 | 62.9 |
| SSSS [12] CVPR'20 | WideResNet-38 | $\mathcal{I}$ | ✓ | 62.7 | 64.3 |
| AA&LR [63] MM'21 | WideResNet-38 | $\mathcal{I}$ | ✓ | 63.9 | 64.8 |
| Ours | WideResNet-38 | $\mathcal{I}$ | ✗ | 65.2 | 65.6 |
| | | | ✓ | **66.4** | **66.8** |

**Semantics-augmented pixel refinement.** To diversify the receptive field, we adopt multiple $3 \times 3$ windows with the set of radiuses $R_2 = \{1, 2, 4, 8, 12, 24\}$ and merge the results. We set the total number of refinement $T = 10$. To generate the pseudo mask $\tilde{M}$ from the refinement score map $M_T$, we use the adaptive threshold $\theta$, *i.e.*, 60% of the maximum scores among all positions for all classes. Also, we ignore conflicting pixels and less confident pixels with the lower bound of 0.2 during the learning from pseudo masks. The balancing parameter $\alpha$ is set to 0.8 by default.

**Region-wise Loss.** To build the boundary and non-boundary sets, we use a local window with its radius $R_3 = 3$. The margin value, *i.e.*, $\gamma$, is set to 3.0.

### 4.2. Comparison with State-of-the-arts

#### 4.2.1. Single-stage results

In Table 1, we compare our model with existing state-of-the-art methods in terms of mIoU on PASCAL VOC 2012. For reference, we include fully supervised methods and multi-stage approaches that are not directly comparable to ours. On both validation and test sets, our method achieves a new state-of-the-art performance with large margins of 2.5 % and 2.0 %, respectively. Notably, even without CRF post-processing, our model is able to surpass all the existing single-stage competitors, which clearly manifests the superiority of ours. Moreover, our model shows a comparable performance to several multi-stage counterparts that rely on complicated and costly training.

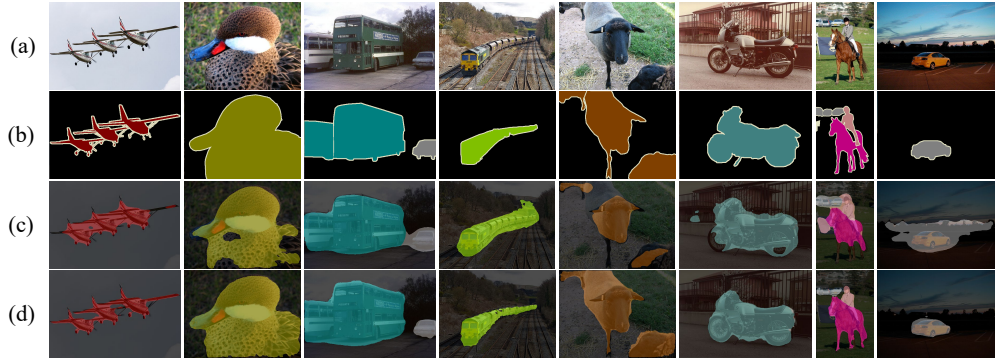

**Figure 5:** Qualitative comparison w/o CRF on PASCAL VOC 2012 validation set. We provide (a) input images, (b) ground-truths, (c) results of SSSS [12], and (d) results of our method.

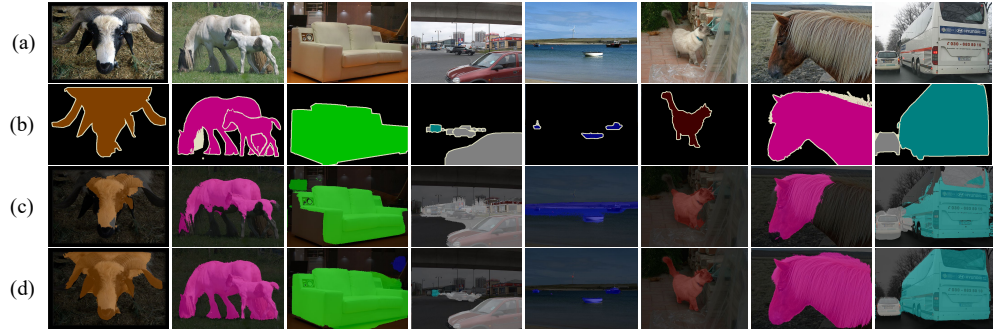

**Figure 6:** Qualitative comparison w/ CRF on PASCAL VOC 2012 validation set. We provide (a) input images, (b) ground-truths, (c) results of SSSS [12], and (d) results of our method.

### 4.2.2. Multi-stage results

To further validate the effectiveness of our model, we adopt it into a simple two-stage framework. Specifically, we first train our model with image-level labels in the single-stage setting. When the training finishes, we feed all training images into the model to generate pseudo masks. Thereafter, we perform CRF on the pseudo masks for offline refinement and utilize the refined pseudo masks as *full supervision* to train an external segmentation model. For the segmentation model, we employ DeepLab based on WideResNet-38 for a fair comparison with existing methods [51, 52].

As shown in Table 1, the simple adoption of our method in the two-stage setting leads to a large performance boost, but at a cost. In the comparison, it outperforms all the previous multi-stage state-of-the-arts under image-level supervision. Furthermore, it even performs favorably against the several recent methods that utilize extra saliency maps. This verifies that our method is capable of generating high-quality pseudo masks.

**Table 2**
Boundary IoU comparison with state-of-the-art approaches on PASCAL VOC 2012 validation set. We reproduce the comparative methods using their official code.

| Method | Sup. | w/o CRF | w/ CRF |
|---|---|---|---|
| DeepLabV3 [5] | $\mathcal{F}$ | 60.4 | 62.8 |
| RRM [19] | $\mathcal{I}$ | 42.0 | 45.4 |
| SSSS [12] | $\mathcal{I}$ | 39.6 | 46.2 |
| Ours | $\mathcal{I}$ | **46.6** | **49.3** |

It should be noted that this paper focuses on the single-stage pipeline. Therefore we conduct the following experiments and analyses in the cost-effective single-stage setting.

### 4.2.3. Qualitative comparison

To better understand the advantages of our model, we perform a qualitative comparison with one of the state-of-the-art methods, SSSS [12]. For a clear comparison, we present both the results w/o and w/ CRF of the comparative methods, which are respectively shown in Fig. 5 and Fig. 6. In the both comparisons w/o and w/ CRF, it is obviously shown that our method produces more precise segmentation masks compared to SSSS. In particular, our results align well with the ground-truth masks regarding the object boundaries, even without the help of CRF as shown in the examples of Fig. 5. This clearly verifies the benefits of our novel shape-injecting strategy.

**Table 3**
Class-wise IoU results on PASCAL VOC 2012 validation set. All the results of the previous methods are the scores after CRF.

| Method | bg | aero | bike | bird | boat | bot. | bus | car | cat | chair | cow | table | dog | horse | mbk | per. | plant | sheep | soft | train | tv | mIoU |
|---|---|---|---|---|---|---|---|---|---|---|---|---|---|---|---|---|---|---|---|---|---|---|
| CRF-RNN [62] | 85.8 | 65.2 | 29.4 | 63.8 | 31.2 | 37.2 | 69.6 | 64.3 | 76.2 | 21.4 | 56.3 | 29.8 | 68.2 | 60.6 | 66.2 | 55.8 | 30.8 | 66.1 | 34.9 | 48.8 | 47.1 | 52.8 |
| RRM [19] | 87.9 | 75.9 | 31.7 | 78.3 | 54.6 | 62.2 | 80.5 | 73.7 | 71.2 | 30.5 | 67.4 | 40.9 | 71.8 | 66.2 | 70.3 | 72.6 | 49.0 | 70.7 | 38.4 | 62.7 | 58.4 | 62.6 |
| SSSS [12] | 88.7 | 70.4 | 35.1 | 75.7 | 51.9 | 65.8 | 71.9 | 64.2 | 81.1 | 30.8 | 73.3 | 28.1 | 81.6 | 69.1 | 62.6 | 74.8 | 48.6 | 71.0 | 40.1 | 68.5 | 64.3 | 62.7 |
| AA&LR [63] | 88.4 | 76.3 | 33.8 | 79.9 | 34.2 | 68.2 | 75.8 | 74.8 | 82.0 | 31.8 | 68.7 | 47.4 | 79.1 | 68.5 | 71.4 | 80.0 | 50.3 | 76.5 | 43.0 | 55.5 | 58.5 | 63.9 |
| Ours w/o CRF | 89.9 | 75.6 | 21.5 | 78.4 | 64.2 | 65.6 | 80.0 | 74.0 | 86.0 | 30.1 | **73.9** | 43.8 | 82.2 | 74.2 | 64.0 | 70.0 | 43.8 | 78.5 | 39.1 | 73.3 | 61.9 | 65.2 |
| Ours w/ CRF | **90.8** | **79.5** | 20.4 | **81.7** | **67.3** | 64.1 | 79.7 | 74.0 | **87.0** | 30.6 | 73.3 | 43.8 | **84.4** | **75.5** | 62.5 | 73.3 | 43.6 | **82.3** | 40.2 | **73.8** | **65.9** | **66.4** |

**Table 4**
Class-wise IoU results on PASCAL VOC 2012 test set. All the results of the previous methods are the scores after CRF.

| Method | bg | aero | bike | bird | boat | bot. | bus | car | cat | chair | cow | table | dog | horse | mbk | per. | plant | sheep | soft | train | tv | mIoU |
|---|---|---|---|---|---|---|---|---|---|---|---|---|---|---|---|---|---|---|---|---|---|---|
| CRF-RNN [62] | 85.7 | 58.8 | 30.5 | 67.6 | 24.7 | 44.7 | 74.8 | 61.8 | 73.7 | 22.9 | 57.4 | 27.5 | 71.3 | 64.8 | 72.4 | 57.3 | 37.3 | 60.4 | 42.8 | 42.2 | 50.6 | 53.7 |
| SSSS [12] | 89.2 | 73.4 | 37.3 | 68.3 | 45.8 | 68.0 | 72.7 | 64.1 | 74.1 | 32.9 | 74.9 | 39.2 | 81.3 | 74.6 | 72.6 | 75.4 | 58.1 | 71.0 | 48.7 | 67.7 | 60.1 | 64.3 |
| Ours w/o CRF | 90.2 | 77.7 | 22.2 | 73.7 | 55.6 | 65.1 | 81.3 | 77.7 | 83.9 | 28.7 | 74.3 | 49.2 | 80.1 | 78.8 | 72.0 | 68.9 | 45.1 | 77.0 | 44.7 | 74.0 | 56.9 | 65.6 |
| Ours w/ CRF | **91.1** | **81.3** | 20.8 | **76.2** | **59.2** | 64.8 | **81.8** | **78.1** | **85.0** | 29.7 | **75.3** | **47.7** | **82.5** | **82.2** | 70.5 | 72.6 | 41.6 | **82.6** | 45.3 | **75.5** | 59.4 | **66.8** |

**Table 5**
Effect of each component in our model. SCM: shape cue module. SPR: semantics-augmented pixel refinement. bIoU stands for the boundary IoU.

| Variants | Baseline | SCM | SPR | mIoU (%) | | bIoU (%) | |
|---|---|---|---|---|---|---|---|
| | | | | *train* | *val* | *train* | *val* |
| #1 | ✓ | | | 42.7 | 41.7 | 18.2 | 17.8 |
| #2 | ✓ | | ✓ | 45.5 | 44.1 | 23.1 | 22.2 |
| #3 | ✓ | ✓ | | 53.0 | 51.0 | 28.3 | 27.8 |
| #4 | ✓ | ✓ | ✓ | **68.9** | **65.2** | **48.9** | **46.6** |

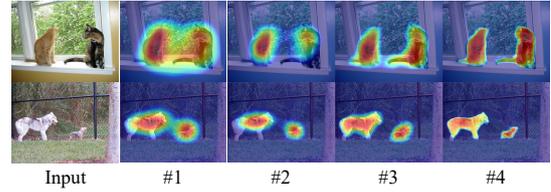

**Figure 7:** Qualitative comparison of the ablated variants.

### 4.2.4. Boundary evaluation

To examine the effectiveness of our method in boundary alignment, we further evaluate it using boundary IoU [77]. The metric has recently been proposed to measure the boundary quality of segmentation masks by computing IoUs only for pixels within distance $d$ from object outlines. We set $d$ to 5 % of an image diagonal, resulting in 30-pixel distances on average.

The results are shown in Table 2, where our method achieves a new state-of-the-art performance with a considerable gap without regard to the use of CRF post-processing. In addition, our method w/o CRF even surpasses the existing state-of-the-arts w/ CRF, which is consistent with the shape-aligned segmentation results shown in the qualitative comparison (Fig. 6). By combining these comparison results on boundary quality with those on the conventional evaluation metric (*i.e.*, mIoU), the superiority of our method over previous state-of-the-arts is clearly validated.

### 4.2.5. Class-wise segmentation results

The class-wise IoUs on the PASCAL VOC 2012 validation and test sets are respectively presented in Table 3 and Table 4. On both sets, our model outperforms the state-of-the-art methods for the majority of the object classes, which clearly demonstrates its superiority. In addition, even without using CRF (Ours w/o CRF), our method beats all the competitors for many object classes as well as in terms of the overall mIoU.

### 4.3. Ablation Study and Analysis
#### 4.3.1. Effects of the proposed components

We investigate the contribution of each proposed component upon the baseline. Here the baseline denotes the common segmentation pipeline with the conventional color-based refinement [19, 12]. We measure mIoU and boundary IoU on PASCAL VOC 2012 *train* and *val* sets. The results are presented in Table 5. On one hand, SPR moderately improves the performance (#1→#2), suggesting that taking class semantics into account during mask refinement is helpful. On the other hand, adding SCM leads to a larger performance gain (#1→#3), which exhibits the importance of shape cues for segmentation. It is also noticed that, the boundary quality is greatly improved thanks to the shape cues, as expected. With both modules put together (#4), our model achieves significant performance boosts compared to the cases using either of them, indicating their synergic property. Intuitively, leveraging shape cues during decoding brings about better initial predictions, which subsequently aid SPR in generating more accurate pseudo masks.

In Fig. 7, we qualitatively compare the ablated variants by visualizing their individual CAMs. We observe that the baseline (#1) produces rough and inaccurate activation maps. In addition, SPR (#2) improves the overall activation map quality, while SCM (#3) helps to produce boundary-aligned activations. Equipped with both the modules, our full model (#4) predicts the activation maps that best suit the whole object regions.

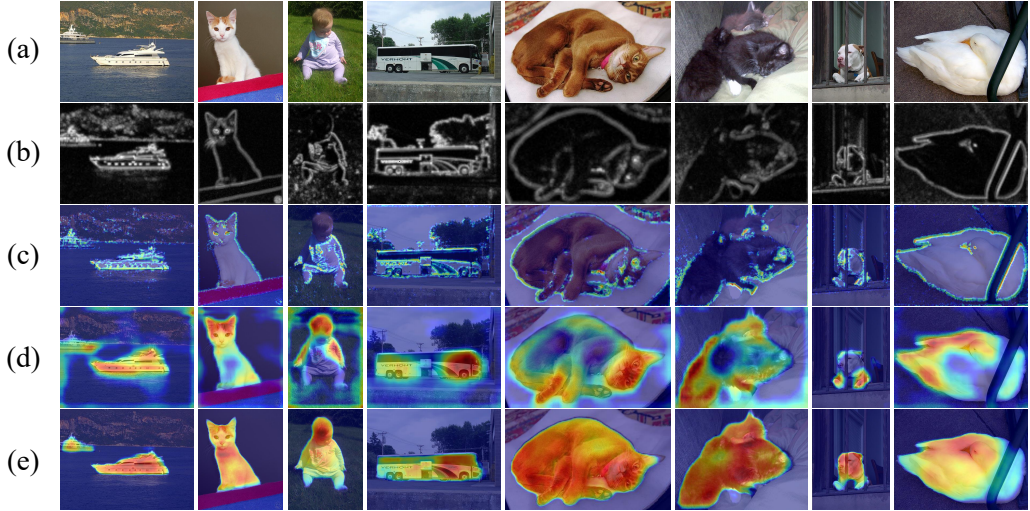

**Figure 8:** Visualization of the effect of SCM. We provide (a) input images, (b) self-information maps, as well as the Grad-CAMs of (c) SCM, (d) the encoder, and (e) the decoder.

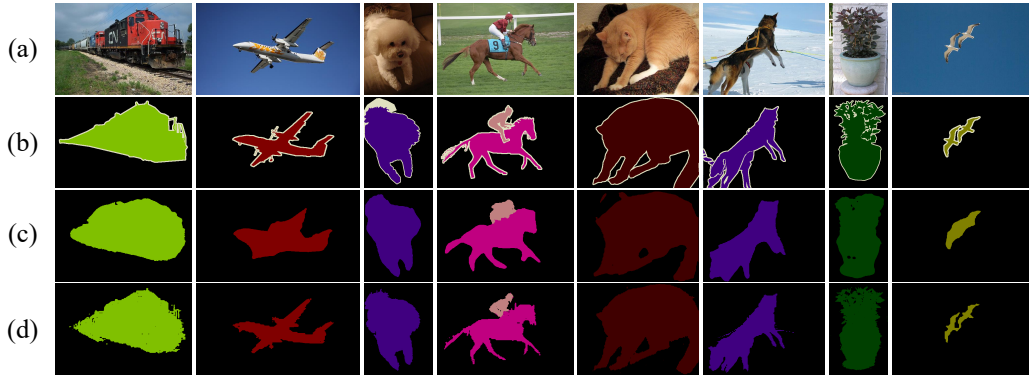

**Figure 9:** Visualization of the effect of SPR. We provide (a) input images, (b) ground-truths, (c) initial predictions, and (d) pseudo masks generated by SPR.

**Table 6**
Analysis on the balancing hyper-parameter $\alpha$. We measure the mIoUs of pseudo masks and predictions for training and validation sets, respectively.

| $\alpha \longrightarrow$ | 0 | $\cdots$ | 0.5 | 0.6 | 0.7 | 0.8 | 0.9 | 1.0 |
|---|---|---|---|---|---|---|---|---|
| train-pseudo | 40.4 | $\cdots$ | 59.9 | 65.4 | **68.8** | 68.5 | 65.6 | 60.0 |
| val | 39.1 | $\cdots$ | 55.5 | 61.2 | 64.5 | **65.8** | 64.8 | 61.9 |

### 4.3.2. Analysis on semantics-augmented pixel refinement

As mentioned in Sec. 3, our SPR extends the conventional color-based refinement [19, 12] by incorporating class semantics as well as color information for mask refinement. More specifically, when $\alpha$ of Eq. 5 is set to 1, SPR becomes the same with PAMR. However, we argue that only considering RGB intensities is insufficient for mask refinement. To validate this claim, we perform an analysis on the impact of $\alpha$ in Table 6. As a result, our method achieves the best performance when $\alpha = 0.8$, with a large gain of 3.9 % from the color-based refinement ($\alpha = 1$). It confirms that both color and class affinities are important for obtaining accurate pseudo masks, and balancing between them is necessary.

### 4.3.3. Visualization of the effect of shape cue module

To better understand the effect of our shape cue module (SCM), we provide visualization results in Fig. 8. From the examples, we make several observations: (1) The self-information maps (Fig. 8b) capture well the shapes (or edges) despite some noises in the background or complicated texture regions. (2) Our SCM (Fig. 8c) extracts useful shape-related features that are sensitive to object silhouettes while filtering out the noises. (3) Guided by the shape cues from SCM, the decoder (Fig. 8e) produces more precise activation maps that are well-aligned with object shapes, compared to the encoder (Fig. 8d). More specifically, our SCM enables sharpening the coarse activation map ($1^{st}$ column) by capturing comprehensive object masks rather than small discriminative parts ($2^{nd}$-$6^{th}$ columns) and producing accurate activation maps even for occluded cases ($7^{th}$-$8^{th}$ columns). This clearly manifests the effectiveness of our shape cues.

**Table 7**
Training time comparison with multi-stage approaches on PASCAL VOC 2012 training set. We reproduce the comparative methods using their official code.

| Method | Stage | CAM training | Multi-scale CAM generation | Auxiliary model training | Pseudo label generation | Segmentation model training | Total time |
|---|---|---|---|---|---|---|---|
| AffinityNet [10] | Multi | 15 min | 12 min | 113 min | 4 min | 340 min | 484 min |
| IRN [10] | Multi | 15 min | 12 min | 21 min | 33 min | 340 min | 421 min |
| Ours | Single | - | - | - | - | 372 min | 372 min |

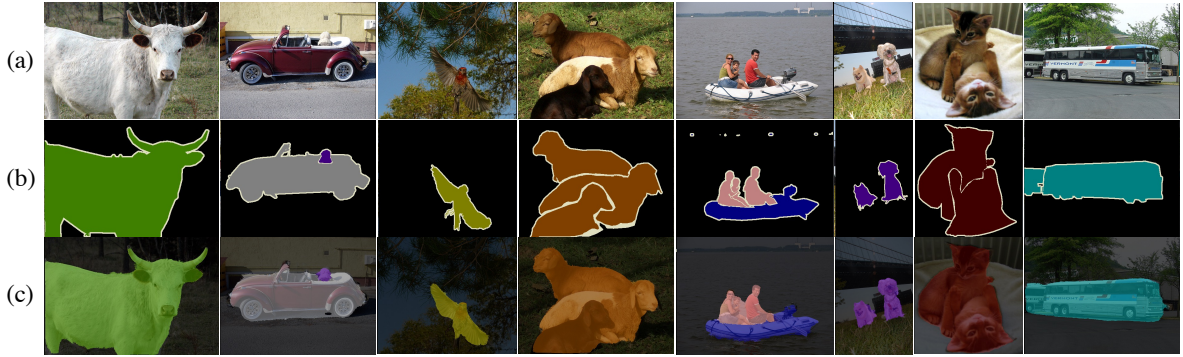

**Figure 10:** Qualitative results on PASCAL VOC 2012 training set. We provide (a) input images, (b) ground-truths, (c) results of our method w/ CRF.

**Table 8**
Effects of loss functions. "bIoU" indicates the boundary IoU.

| $\mathcal{L}_{cls}$ | $\mathcal{L}_{pixel}$ | $\mathcal{L}_{region}$ | mIoU (%) train | mIoU (%) val | bIoU (%) train | bIoU (%) val |
|---|---|---|---|---|---|---|
| ✓ | | | 35.7 | 35.6 | 22.4 | 22.0 |
| ✓ | ✓ | | 61.6 | 58.1 | 38.1 | 36.4 |
| ✓ | | ✓ | 33.7 | 33.5 | 23.0 | 22.7 |
| ✓ | ✓ | ✓ | **68.9** | **65.2** | **48.9** | **46.6** |

#### 4.3.4. Visualization of the effect of semantics-augmented pixel refinement

To see the effect of our mask refinement (*i.e.*, SPR), we visualize several examples in Fig. 9. As shown in the figure, initial predictions (Fig. 9c) are already accurate and comprehensive to a certain extent thanks to our shape cues, but the boundaries are somewhat rough and less match the ground truths due to the large receptive field. After performing our SPR on them, we obtain the more accurate pseudo masks with well-aligned outlines (Fig. 9d). They are in turn used to effectively guide our model in the self-supervised training stage.

One limitation of our SPR could be the inflexibility in balancing between the two types of information. Naturally, different object classes may have their own optimal balancing weights for refining their masks, and it may hold true even for the instances within the same class. This makes our SPR produce potentially sub-optimal pseudo masks, which in turn restricts the improvements of self-supervised training of the segmentation method. Designing adaptable (or learnable) balancing weights could be a promising direction for future work.

#### 4.3.5. Effects of the loss functions

We perform ablation studies on the loss functions to inspect their contributions in Table 8. When no pseudo mask is used ($\mathcal{L}_{cls}$ only), our model shows the inferior performance. When the pixel-wise loss ($\mathcal{L}_{pixel}$) is adopted, the performance significantly increases, indicating the important role of our SPR module. On the other hand, solely using the region-wise loss ($\mathcal{L}_{region}$) does not bring a performance gain. We conjecture that the region-wise loss could be misleading without pixel-level constraints, since the boundary sets from pseudo masks are unreliable, especially at the early steps. Meanwhile, the two losses from pseudo masks play complementary roles, and our model attains the best performance when they are used together.

#### 4.3.6. Training cost analysis

To verify the training efficiency of our method over the existing multi-stage approaches, we conduct training cost analysis. For the competitors, we choose AffinityNet [10] and IRN [18], since they are widely used as baselines for the latest two-stage works such as SEAM [11], AdvCAM [16], and RIB [17], *i.e.*, they require at least the training time of AffinityNet and IRN. The experiments are conducted using two GTX 1080Ti GPUs and the Intel Core i7-8700 Processor. The results can be found in Table 7. It can be observed that our method is much more training-efficient compared to the multi-stage approaches, as it can be trained in a single-stage manner without relying on auxiliary segmentation networks. In addition, our model does not require complicated training strategies including expensive multi-scale CAM generation and carefully designed auxiliary models. Moreover, it can also be noticeable that our method puts only a minor computational overhead (about 10%) upon the baseline segmentation model, *i.e.*, DeepLabV3 [5].

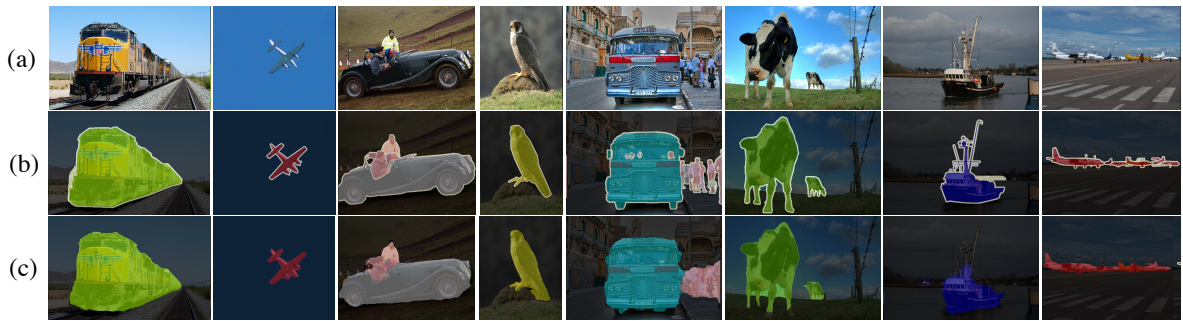

**Figure 11:** Qualitative results on PASCAL VOC 2012 validation set. We provide (a) input images, (b) ground-truths, (c) results of our method w/ CRF.

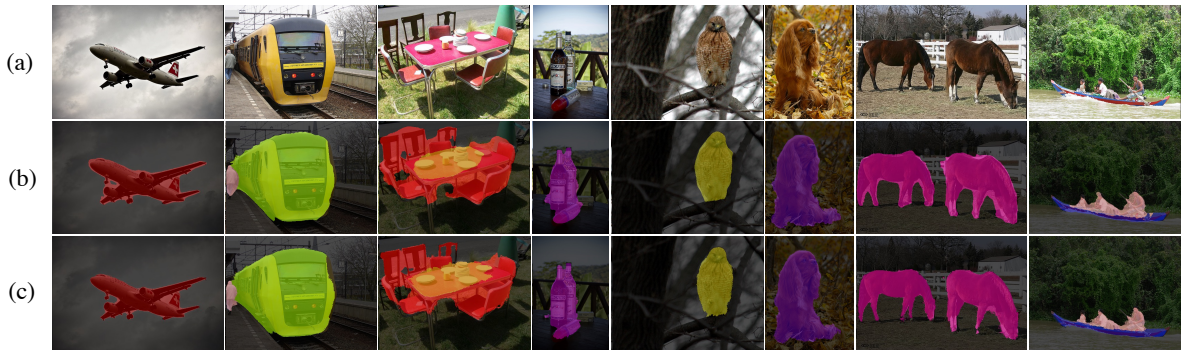

**Figure 12:** Qualitative results on PASCAL VOC 2012 test set. We provide (a) input images, (b) results of our method w/o CRF, (c) results of our method w/ CRF.

### 4.4. Qualitative Results

We provide more visualization results to further demonstrate the strong performance of our model. The qualitative results on the PASCAL VOC 2012 training, validation, and test sets are respectively presented in Fig. 10, Fig. 11, and Fig. 12. For the training and validation sets, we show the ground-truth images and the results after CRF. On the other hand, we instead demonstrate the results of our method w/o and w/ CRF for the test set, as the ground-truths of the test split are withheld. As can be seen in Fig. 10 and Fig. 11, our method is able to produce very precise segmentation maps that well match the ground-truths even for the challenging cases (*e.g.*, complex background, cluttered objects). Moreover, in Fig. 12, it is shown that our method produces comprehensive segmentation predictions with well-aligned boundaries even without CRF. After CRF post-processing, the results are further improved and show the high segmentation quality, even though the model is trained using only image-level labels.

### 5. Conclusion and Discussion

In this paper, we presented a novel framework for weakly supervised semantic segmentation. We started by associating the locality of CAMs with the texture bias of CNNs. To handle it and produce comprehensive segmentation masks, we proposed to extract shape information from the encoder and explicitly use it as *shape cues* for segmentation. Moreover, we designed the semantics-augmented pixel refinement in which pseudo masks are obtained using pixel-wise affinities that consider class semantics as well as color intensity. By the in-depth analyses, we verified the efficacy and the complementarity of the proposed methods. Furthermore, we achieved a new state-of-the-art on *val* and *test* sets of PASCAL VOC 2012 in both single and multi-stage settings.

Our proposed model effectively leverages the shape cue to produce the segmentation results aligned with the object boundaries, greatly improving the performance in the weakly-supervised setting. On the other hand, the performance of our model may depend on the quality of shape information to an extent, e.g., noisy shape information can mislead the model. In the future work, we would like to explore Vision Transformers [78] for obtaining shape information which are recently found to focus more on the object shapes compared to CNN variants [79].

### Acknowledgement

This project was partly supported by the National Research Foundation of Korea grant funded by the Korea government (MSIT) (No. 2022R1A2B5B02001467) and the Institute for Information & Communications Technology Planning & Evaluation (IITP) grant funded by the Korea government (No. 2020-0-01361: Artificial Intelligence Graduate School Program (YONSEI UNIVERSITY)).

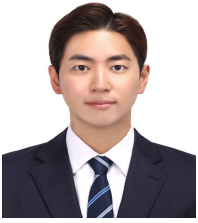

**Sungpil Kho** received the B.S double degree in Advanced Material Engineering and Computer Engineering from Kyunghee University. He is currently a M.S student at the Graduate School of Computer Science, Yonsei University, Seoul, Korea. His research interests include computer vision, deep learning and machine learning, especially semantic segmentation.

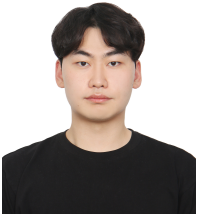

**Pilhyeon Lee** is currently a Ph.D. student in Computer Science at Yonsei University, Seoul, Korea. He received the B.S. degree in Computer Science and Engineering from Chung-Ang University, Seoul, Korea. His research interests include video analysis, weakly-supervised learning, and representation learning.

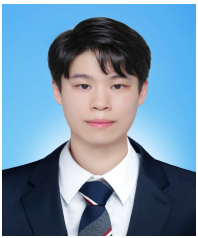

**Wonyoung Lee** received the B.S. degree in Computer Engineering from Inha University. He is currently a M.S student at the Graduate School of Artificial Intelligence, Yonsei University, Seoul, Korea. His research interests include computer vision, deep learning and machine learning.

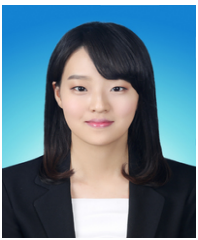

**Minsong Ki** received the B.S. degree in Computer Science from Duksung Women's University and Ph.D. degree in Computer Science from Yonsei University. She is currently an AI Researcher at LG Uplus. Her research interests include weakly supervised learning, object localization, detection, face recognition, and deep learning.

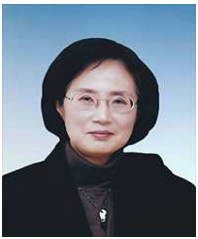

**Hyeran Byun** is currently a professor of Computer Science at Yonsei University. She was an Assistant Professor at Hallym University, Chooncheon, Korea, from 1994 to 1995. She served as a non-executive director of National IT Industry Promotion Agency (NIPA) from Mar. 2014 to Feb. 2018. She is a member of National Academy Engineering of Korea. Her research interests include computer vision, image and video processing, deep learning, artificial intelligence, machine learning, and pattern recognition. She received the B.S. and M.S. degrees in mathematics from Yonsei University, Seoul, Korea, and the Ph.D. degree in computer science from Purdue University, West Lafayette, IN, USA.